\documentclass[journal]{IEEEtran}
\usepackage{bibentry}
\usepackage{xcolor,soul,framed} 

\colorlet{shadecolor}{yellow}
\usepackage[pdftex]{graphicx}
\graphicspath{{../pdf/}{../jpeg/}}
\DeclareGraphicsExtensions{.pdf,.jpeg,.png}

\usepackage[cmex10]{amsmath}
\usepackage{array}
\usepackage{mdwmath}
\usepackage{mdwtab}
\usepackage{eqparbox}
\usepackage{url}
\usepackage{multirow}
\usepackage{comment}
\usepackage{amssymb}
\usepackage{algorithmic}
\usepackage{algorithm}

\usepackage{xcolor}
\usepackage{ulem}

\begin{document}
\bstctlcite{IEEEexample:BSTcontrol}
\title{Enhancing the Reliability of Segment Anything Model for Auto-Prompting Medical Image Segmentation with Uncertainty Rectification}
\author{Yichi Zhang, Shiyao Hu, Sijie Ren, Chen Jiang, Yuan Cheng and Yuan Qi
	\thanks{Yichi Zhang, Sijie Ren, Yuan Cheng and Yuan Qi is with Artificial Intelligence Innovation and Incubation Institute, Fudan University, Shanghai, China, and with Shanghai Academy of Artificial Intelligence for Science, Shanghai, China. Corresponding authors: Yuan Cheng and Yuan Qi. (cheng\_yuan@fudan.edu.cn, qiyuan@fudan.edu.cn)}
	\thanks{Shiyao Hu is with School of Computer Science and Technology, Xi’an Jiaotong University, Xi'an, China.}
	\thanks{The computations in this research were performed using the CFFF platform of Fudan University.}
	
}
\maketitle

\begin{abstract}
The Segment Anything Model (SAM) has recently emerged as a groundbreaking foundation model for prompt-driven image segmentation tasks.
However, both the original SAM and its medical variants require slice-by-slice manual prompting of target structures, which directly increase the burden for applications. Despite attempts of auto-prompting to turn SAM into a fully automatic manner, it still exhibits subpar performance and lacks of reliability especially in the field of medical imaging.
In this paper, we propose UR-SAM, an uncertainty rectified SAM framework to enhance the reliability for auto-prompting medical image segmentation. Building upon a localization framework for automatic prompt generation, our method incorporates a prompt augmentation module to obtain a series of input prompts for SAM for uncertainty estimation and an uncertainty-based rectification module to further utilize the distribution of estimated uncertainty to improve the segmentation performance.
Extensive experiments on two public 3D medical datasets covering the segmentation of 35 organs demonstrate that without supplementary training or fine-tuning, our method further improves the segmentation performance with up to 10.7 \% and 13.8 \% in dice similarity coefficient, demonstrating efficiency and broad capabilities for medical image segmentation without manual prompting.

\end{abstract}


\IEEEpeerreviewmaketitle

\section{Introduction}
\IEEEPARstart{M}{edical} image segmentation aims to delineate the interested anatomical structures like organs and tumors from the original images by labeling each pixel into a certain class, which is one of the most representative and comprehensive research topics in both communities of computer vision and medical image analysis \cite{MIA2017survey,Lynch2018NewMT}. 
Accurate segmentation can provide reliable volumetric and shape information of target structures, so as to assist in many further clinical applications like disease diagnosis, quantitative analysis, and surgical planning \cite{bernard2018deep,heller2020state,lalande2021deep,AbdomenCT-1K}. Since manual contour delineation is labor-intensive and time-consuming and suffers from inter-observer variability, it is highly desired in clinical studies to develop automatic medical image segmentation methods.
With the unprecedented developments in deep learning, deep neural networks have been widely applied and achieved great success in the field of medical image segmentation due to their outstanding performance \cite{ronneberger2015u,isensee2020nnunet}.
However, existing deep models are often tailored for specific modalities and targets, which limits their capacity for further generalization. 
The recent introduction of the Segment Anything Model (SAM) \cite{SAM} has gained massive attention as a promptable foundation model capable of generating fine-grade segmentation masks using prompts like points or bounding boxes, demonstrating impressive performance on a variety of semantic segmentation tasks \cite{semanticSAM,trackanything}.
However, recent studies have revealed SAM’s limited performance in specific domain tasks \cite{ji2023segment}, such as medical image segmentation where challenges emerge in scenarios characterized by high structural complexity and low contrast, leading to weak boundaries \cite{SAM4MIS,SAM-SZU, SAM-Empirical}.
Besides, most of these applications to medical image segmentation require manual prompting of target structures to obtain acceptable performance, which is still labor-intensive. 
Despite attempts of auto-prompting to turn SAM into a fully automatic manner \cite{MedLSAM}, it still exhibits subpar performance and lacks reliability, while guaranteeing the reliability of segmentation results is of great importance, especially for medical imaging where the variability in segmentation accuracy directly contributes to safeguarding patients' safety during clinical procedures.
One promising avenue to issue this challenge is uncertainty estimation, which serves as a valuable approach to provide the reliability of medical image segmentation since it allows us to quantify the confidence of the model’s output and identify when the model may not perform well, which has demonstrated its reliability and robustness in many medical image segmentation tasks \cite{zou2022tbrats, zhang2023uncertainty}.
In this paper, we propose UR-SAM, an uncertainty rectified SAM framework to enhance the reliability for auto-prompting medical image segmentation by estimating the segmentation uncertainty and utilizing uncertainty for rectification to enhance the reliability and improve the accuracy of SAM for medical image segmentation.
Since different prompts may yield diverse results, instead of adding perturbations to input images or model parameters, we focus on prompt augmentation to introduce perturbations and obtain a series of different segmentation outputs. Then we establish pixel-level confidence evaluation through uncertainty estimation based on these results, which can be utilized to identify areas of concern and provide additional information to the clinician along with model-generated predictions.
To further utilize estimated uncertainty and improve the performance, we propose a class-specific confidence-based filtering method to select out high uncertainty regions and an uncertainty rectification module to divide regions within a certain range of image intensity into target areas.

To evaluate the effectiveness of our proposed framework, we conduct experiments based on the original SAM \cite{SAM} and the medical-adapted MedSAM \cite{MedSAM} as the foundation for our framework on two public 3D datasets including the segmentation of 22 head and neck organs and 13 abdominal organs for a comprehensive evaluation.
Our experiments demonstrate significant performance improvement of SAM’s segmentation result and robustness to different prompts.
The main contribution of our work can be summarized as follows:

\begin{itemize}
\item We present \textbf{UR-SAM}, an Uncertainty Rectified Segment Anything Model by incorporating uncertainty estimation and rectification to enhance the robustness and reliability for auto-prompting medical image segmentation.
\item We propose to augment given prompts by adding perturbations with pre-defined ratios to estimate segmentation uncertainty based on the predictions of multi-prompt input, since the segmentation performance of SAM is sensitive to the input prompt. 
\item We propose to utilize estimated segmentation uncertainty with class-specific confidence-based filtering to select out high uncertain regions for rectification to further improve the segmentation performance. 
\end{itemize}

\begin{figure*}[t]
	\includegraphics[width=18cm]{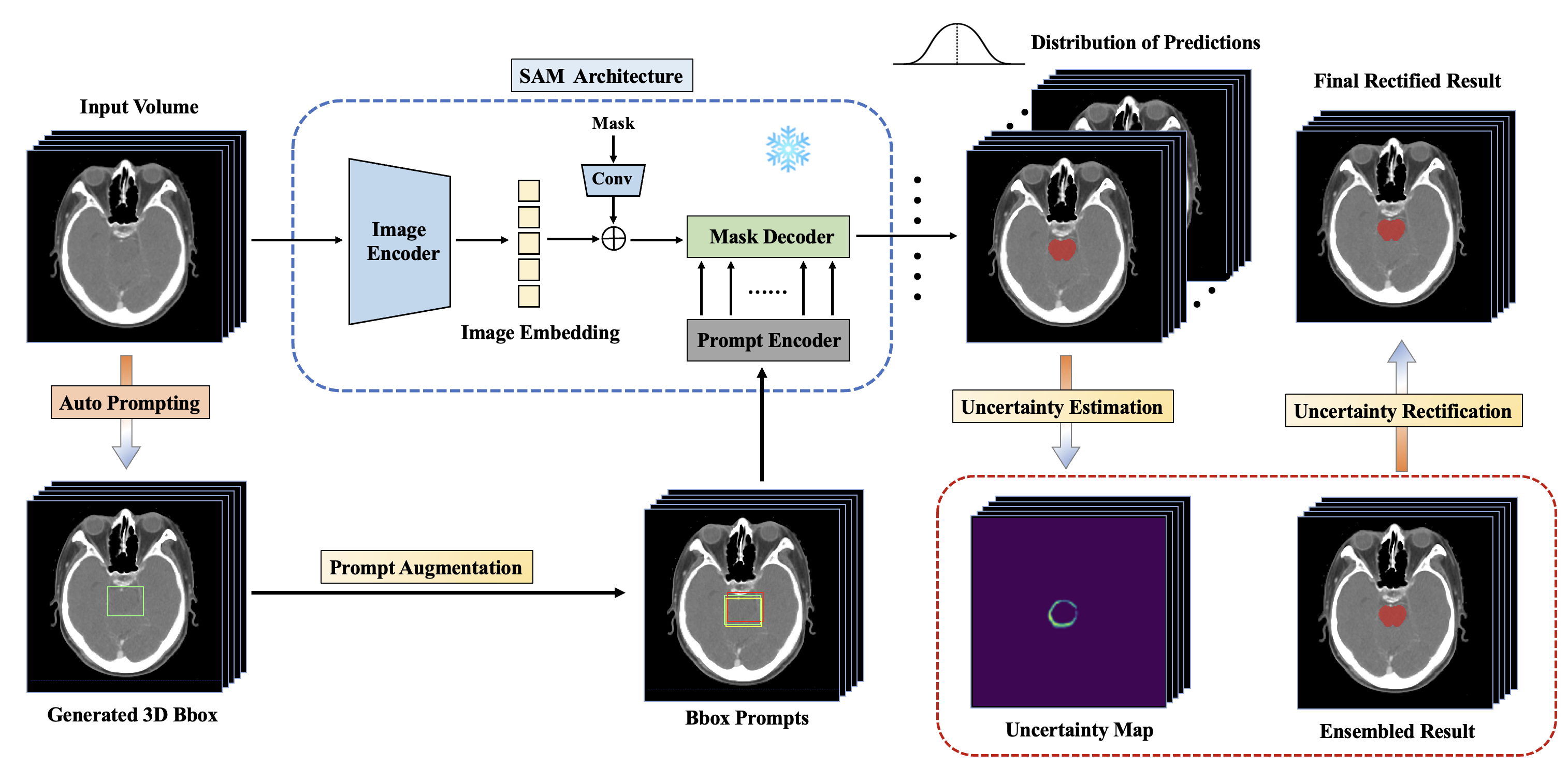}
	\caption{Overview of our proposed Uncertainty Rectified Segment Anything Model (UR-SAM) framework. We utilize an landmark localization model for auto-prompting with augmentation to generate a series of different bounding box prompts $B$ for each image $x$. we can obtain the distributions of predictions for uncertainty estimation. The SAM architecture in red-line utilizes the original image and generated prompts to obtain a set of predictions for uncertainty estimation.
 Then the estimated uncertainty can be utilized for rectification to obtain the final segmentation results.}
	\label{Architecture}
\end{figure*}

\section{Related Work}

\subsection{Foundation Models}

Foundation models are a rapidly expanding field in artificial intelligence research, focus on developing large-scale, general-purpose language and vision models. These models are often trained on large-scale datasets, enabling them to acquire general representations and capabilities that can be applied across different domains and applications. The GPT (Generative Pre-trained Transformer) series \cite{GPT-3,GPT-4} is one of the most widely known foundation models, which demonstrated impressive capabilities and outstanding performance in various natural language processing tasks.
The success of these foundation models has inspired researchers to develop large-scale vision foundational models, with a focus on capturing cross-modal interactions between vision and language \cite{CLIP,ALIGN,DALL-E}.  
Foundation models have also demonstrated strong potential in addressing a wide range of downstream tasks for medical image analysis, accelerating the development of accurate and robust models~\cite{zhang2023challenges}.

\subsection{Segment Anything Model for Medical Images}

As the first promptable foundation model for segmentation tasks, the Segment Anything Model (SAM) \cite{SAM} is trained on the large-scale SA-1B dataset with an unprecedented number of images and has shown strong zero-shot generalization for natural image segmentation. 
As a very important branch of image segmentation, recent studies have explored the application of SAM to medical image segmentation \cite{SAM4MIS} like benchmarking SAM on different medical image segmentation tasks including pathology segmentation \cite{SAM-pathology}, surgical instrument segmentation \cite{SAM-RS}, CT images \cite{SAM-LiverTumor,SAM-DKFZ-Abdomen}, MRI images \cite{SAM-BrainMR}, and several multi-modal, multi-dataset evaluations \cite{SAM-Meds,SAM-Empirical,SAM-MI,SAM-SZU}.
These evaluation results on different datasets have shown that SAM has limited generalization ability when directly applied to medical image segmentation, which varies significantly across different datasets and tasks.
To better adapt SAM for medical images, several studies focus on fine-tuning SAM on medical datasets \cite{MedSAM,Med-SAM-Adapter,SAMed,3DSAM-adapter}, auto-prompting \cite{MedLSAM,AP-SAM,autosam} and assisting in other segmentation networks \cite{zhang2023semisam,li2023nnsam,li2023segment} to better adapt SAM to medical image segmentation.
Although these approaches can improve the unsatisfactory segmentation results to some extent, the performance is still not sufficient for clinical applications where the reliability of segmentation requires further study.

\subsection{Uncertainty Estimation}

In deep learning, uncertainty estimation is an essential task that can lead to significant improvements in the reliability and trustworthiness of deep models, which is particularly important in medical imaging where the uncertainty can be used to identify areas of concern or to provide additional information to the clinician \cite{UncSurvey,MedUncReview}.
The quantification of uncertainty involves two fundamental types: aleatoric uncertainty which represents inherent noise or variation in the data learned by the model, and epistemic uncertainty which quantifies the inherent lack of knowledge about the underlying model architecture and parameters \cite{der2009aleatory}.
For image segmentation, this uncertainty acknowledges that there may be uncertainty in determining whether a pixel belongs to the object or the background. The model assigns a probability value that represents the confidence or uncertainty associated with each pixel's classification, enabling a more nuanced and informative representation of the segmentation task to capture the uncertainty inherent in the data \cite{kendall2017uncertainties}.
By addressing and quantifying these uncertainties, the reliability and accuracy of deep models can be enhanced, making them more suitable for real-world applications.

\section{Methods}

The overall architecture of our proposed framework is shown in Fig. \ref{Architecture}, where we aim to enhance the reliability and improve the accuracy by evaluating and incorporating uncertainty into the segmentation workflow. 
We first use a localization framework for the identification of the extreme points of target organs to generate bounding box prompts for the subsequent segmentation procedure. 
After that, the initial prompt is augmented by adding perturbations to generate a series of slightly different prompts for segmentation.
Since these different prompts may cause variances in the segmentation results, we can approximate the segmentation uncertainty of the model with the predictive entropy.
Finally, the estimated uncertainty is utilized for the rectification of segmentation results through confidence-based filtering to select out high uncertain regions for rectification to further improve the segmentation performance.

\begin{figure*}[t]
	\includegraphics[width=18cm]{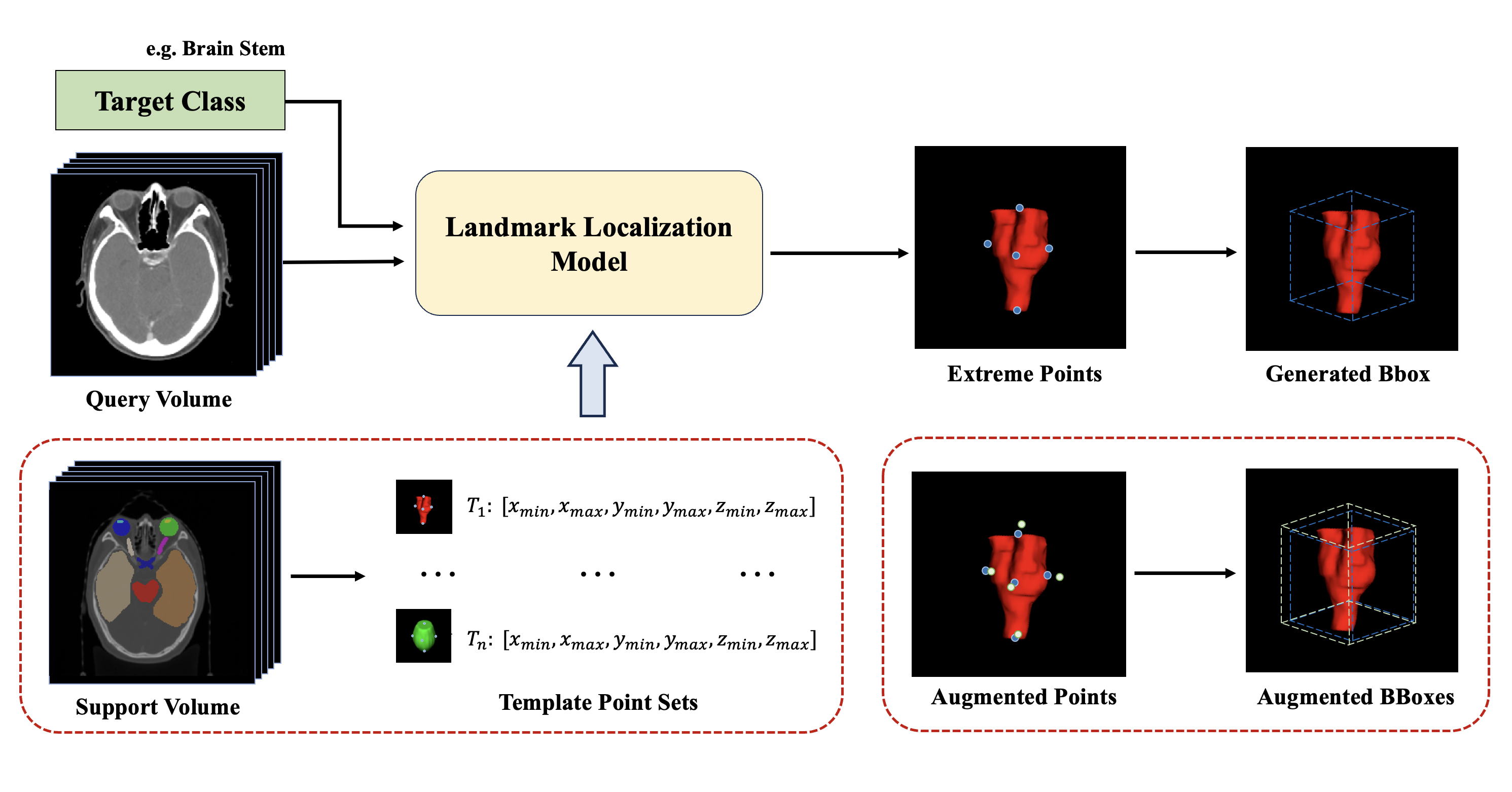}
	\caption{Detailed pipeline of prompt generation in our framework. For each target organ, six extreme points in three volumetric directions is localized to generate bounding box prompts. Furthermore, the initial generated bounding box prompt is augmented by random shifting within pre-defined ratios.}
	\label{AutoPrompting}
\end{figure*}

\subsection{Landmark Localization for Auto-Prompting}
\label{Landmark}
As shown in Fig. \ref{AutoPrompting}, we adopt a landmark localization framework to automatically identify the extreme points of target organs from the 3D volumetric medical images following the design in \cite{MedLSAM} for bounding box prompt generation, based on the assumptions that the same anatomical structure in different images corresponds to the same latent coordinates.
The localization model is formed in a two-step approach.
Firstly, a Relative Distance Regression module (RDR) is adopted to extract high-level features and map input images from different individuals onto a unified anatomical coordinate system. This enables coarse localization of the point by predicting the 3D offset between the center points of the query patch and the support patch. 
Given the inherent variations in anatomical positioning across different individuals, regions sharing the same latent coordinates in various images may still correspond to different anatomical structures. Therefore, a Multi-Scale Similarity (MSS) component is adopted to refine the localization by incorporating local pixel-level features from points of interest to identify the most similar feature within the vicinity of the initially localized point.

During the training process, a self-supervised approach is employed, where two patches are randomly selected from the same image to serve as the query patch and support patch, and their augmentations are obtained. Both the RDR and MSS components share a multi-layer encoder, where RDR utilizes Mean Square Error (MSE) loss to regress the displacement between the query and support patches, while MSS treats each patch and its augmentation as a pair of data, using Cross-Entropy (CE) loss to ensure that the features of the patch images and their augmentations at the same location are as similar as possible. This self-supervised training approach allows the model to learn representations that facilitate accurate localization and similarity assessment.
During inference, a small amount of support volumes is utilized as template point sets.
Firstly, the RDR component is utilized to obtain the coarse coordinate position. Subsequently, a patch cropped around the position is used as the query for the MSS module to identify the pixel with the highest similarity to the corresponding pixel, providing the final prediction result.
For each target, after successfully identifying the six extreme points of the target organ, we can get the bounding box of the target as the prompt for the subsequent segmentation procedure.

\subsection{The SAM Architecture}
The overall architecture of SAM \cite{SAM} is a prompt-driven image segmentation architecture known for its impressive performance and generalization ability for image segmentation.
SAM consists of three main components including an image encoder, a prompt encoder, and a mask decoder.
The image encoder uses the Vision Transformer (ViT) \cite{ViT2020} to transform original images into discrete embeddings. The prompt encoder converts diverse prompts including sparse prompts and dense prompts into compact embeddings by combining fixed positional encoding and adaptable prompt-specific embeddings. The mask decoder receives the extracted information from both the image encoder and the prompt encoder and incorporates prompt self-attention and cross-attention in two directions for prompt-to-image and image-to-prompt attention to update the image embeddings and prompt embeddings. The processed feature map is up-sampled and then passes through a multi-layer perception (MLP) to generate segmentation masks.

\subsection{Prompt Augmentation for Uncertainty Estimation}

To enhance the reliability and accuracy of SAM, we aim to evaluate and incorporate uncertainty into the segmentation workflow. 
Instead of adding perturbations to input images or model parameters, we focus on augmenting input prompts for SAM, since slightly different bounding box prompts may cause variances in the segmentation results even when they refer to the same object given the same image.
For model-generated bounding box prompt $b$, we conduct a prompt augmentation procedure to add perturbations to the initial prompt by random shifting to generate augmented bounding box prompts $B=\{b^1,b^2,\cdots,b^n\}$, where $n$ is a pre-defined number for augmentation.
With different prompt initializations, each bounding box prompt guides the model to generate different segmentation results $Y=\{y^1,y^2,\cdots,y^n\}$.
By ensembling these outputs, we can get the ensembled segmentation result $\hat y$ and summarize the predictive entropy to approximate the segmentation uncertainty $u(\hat y)$ of the model as follows:

\begin{equation}
\hat y = \frac{1}{n}\sum\limits_{i = 1}^n y^{i} = \frac{1}{n}\sum\limits_{i = 1}^n {{f_{SAM}}\left( {x,{b^i}} \right)} 
\label{ensemble}
\end{equation}

\begin{equation}
u(\hat y) = -\sum p(y^i|x) \log{p(y^i|x)}
\label{unc}
\end{equation}

\begin{table*}
	\caption{Segmentation performance of Dice Similarity Coefficient (DSC) with different augmentation numbers and perturb ratios for prompt augmentation using MedSAM backbone for 3D head-and-neck organ segmentation in the StructSeg dataset. } \label{PromptAug_StructSeg}
	\centering
	\renewcommand\arraystretch{1.2}
        \scriptsize
\begin{tabular}{p{1cm}|p{0.3cm}p{0.3cm}p{0.3cm}p{0.3cm}p{0.3cm}p{0.3cm}p{0.3cm}p{0.3cm}p{0.3cm}p{0.3cm}p{0.3cm}p{0.3cm}p{0.3cm}p{0.3cm}p{0.3cm}p{0.3cm}p{0.3cm}p{0.3cm}p{0.3cm}p{0.3cm}p{0.4cm}p{0.4cm}|c}
\hline \hline
Model (num/ratio) & BS & E-L & E-R & L-L & L-R & ON-L & ON-R & OC & TL-L & TL-R & P & PG-L & PG-R & IE-L & IE-R & ME-L & ME-R & J-L & J-R & SC & M-L & M-R & Avg \\ \hline
w/o Aug & 0.643 & 0.652 & \textbf{0.664} & 0.280 & 0.264 & 0.385 & \textbf{0.369} & 0.320 & \textbf{0.705} & \textbf{0.733} & 0.404 & \textbf{0.532} & 0.464 & 0.669 & 0.644 & 0.480 & 0.435 & 0.562 & 0.436 & \textbf{0.117} & 0.050 & 0.161 & 0.453 \\ \hline
3 / 0.005 & \textbf{0.686} & 0.695 & 0.663 & \textbf{0.285} & \textbf{0.265} & \textbf{0.396} & 0.365 & \textbf{0.359} & 0.677 & 0.708 & 0.411 & 0.511 & 0.490 & 0.713 & 0.700 & \textbf{0.505} & \textbf{0.498} & \textbf{0.603} & 0.542 & 0.108 & \textbf{0.169} & \textbf{0.170} & 0.477 \\
5 / 0.005 & 0.684 & 0.694 & 0.663 & 0.282 & 0.261 & 0.394 & 0.365 & 0.358 & 0.676 & 0.708 & 0.411 & 0.508 & 0.488 & 0.713 & 0.700 & 0.502 & 0.497 & \textbf{0.603} & 0.544 & 0.108 & 0.168 & 0.169 & 0.477 \\
7 / 0.005 & 0.684 & 0.694 & 0.662 & 0.281 & 0.261 & 0.393 & 0.365 & 0.357 & 0.676 & 0.708 & 0.410 & 0.508 & 0.488 & 0.713 & 0.700 & 0.502 & 0.497 & \textbf{0.603} & 0.544 & 0.108 & 0.168 & \textbf{0.170} & 0.476 \\ \hline
3 / 0.01 & \textbf{0.686} & 0.696 & 0.663 & \textbf{0.285} & \textbf{0.265} & \textbf{0.396} & 0.365 & \textbf{0.359} & 0.677 & 0.708 & 0.411 & 0.511 & \textbf{0.491} & \textbf{0.714} & 0.700 & \textbf{0.505} & \textbf{0.498} & \textbf{0.603} & 0.542 & 0.108 & 0.168 & \textbf{0.170} & \textbf{0.478} \\
5 / 0.01 & 0.684 & 0.694 & 0.663 & 0.282 & 0.261 & 0.394 & 0.366 & 0.358 & 0.675 & 0.707 & 0.411 & 0.509 & 0.487 & 0.713 & 0.700 & 0.502 & \textbf{0.498} & 0.602 & 0.543 & 0.108 & 0.167 & 0.168 & 0.477 \\
7 / 0.01 & 0.684 & 0.694 & 0.662 & 0.281 & 0.261 & 0.393 & 0.365 & 0.357 & 0.675 & 0.706 & 0.410 & 0.507 & 0.487 & 0.713 & 0.700 & 0.502 & 0.497 & \textbf{0.603} & 0.544 & 0.107 & 0.166 & 0.168 & 0.476 \\ \hline
3 / 0.03 & 0.678 & \textbf{0.697} & \textbf{0.664} & 0.284 & 0.262 & 0.384 & 0.368 & 0.355 & 0.671 & 0.704 & 0.417 & 0.500 & 0.477 & \textbf{0.714} & 0.705 & 0.499 & 0.497 & 0.602 & 0.535 & 0.106 & 0.164 & 0.165 & 0.475 \\ 
5 / 0.03 & 0.672 & 0.693 & \textbf{0.664} & 0.281 & 0.258 & 0.382 & 0.372 & 0.350 & 0.668 & 0.699 & \textbf{0.419} & 0.490 & 0.466 & 0.707 & 0.706 & 0.495 & 0.495 & 0.601 & 0.539 & 0.105 & 0.162 & 0.163 & 0.472 \\ 
7 / 0.03 & 0.671 & 0.652 & \textbf{0.664} & 0.280 & 0.264 & 0.385 & \textbf{0.369} & 0.320 & 0.705 & \textbf{0.733} & 0.404 & \textbf{0.532} & 0.464 & 0.669 & 0.644 & 0.480 & 0.435 & 0.562 & 0.436 & \textbf{0.117} & 0.150 & 0.161 & 0.454 \\ \hline
3 / 0.05 &0.669 & 0.693 & 0.659 & 0.282 & 0.255 & 0.385 & 0.367 & 0.352 & 0.663 & 0.698 & 0.418 & 0.490 & 0.465 & \textbf{0.714} & \textbf{0.705} & 0.493 & 0.494 & 0.601 & 0.538 & 0.104 & 0.160 & 0.162 & 0.471 \\ 
5 / 0.05 & 0.655 & 0.683 & 0.657 & 0.277 & 0.249 & 0.384 & \textbf{0.369} & 0.348 & 0.651 & 0.688 & 0.418 & 0.473 & 0.446 & 0.705 & 0.700 & 0.489 & 0.480 & 0.599 & 0.542 & 0.101 & 0.156 & 0.158 & 0.465 \\ 
7 / 0.05 & 0.651 & 0.680 & 0.648 & 0.270 & 0.244 & 0.375 & 0.363 & 0.354 & 0.648 & 0.685 & 0.408 & 0.467 & 0.445 & 0.713 & 0.702 & 0.484 & 0.478 & 0.599 & \textbf{0.548} & 0.101 & 0.154 & 0.157 & 0.462 \\ \hline
3 / 0.1 & 0.643 & 0.672 & 0.646 & 0.270 & 0.241 & 0.369 & 0.365 & 0.348 & 0.638 & 0.677 & 0.418 & 0.463 & 0.436 & 0.707 & 0.697 & 0.478 & 0.478 & 0.588 & 0.536 & 0.098 & 0.150 & 0.153 & 0.458 \\ 
5 / 0.1 & 0.618 & 0.650 & 0.642 & 0.260 & 0.231 & 0.360 & 0.357 & 0.342 & 0.623 & 0.652 & 0.416 & 0.433 & 0.403 & 0.699 & 0.693 & 0.461 & 0.466 & 0.580 & 0.536 & 0.095 & 0.143 & 0.147 & 0.446 \\ 
7 / 0.1 & 0.606 & 0.643 & 0.627 & 0.252 & 0.221 & 0.350 & 0.350 & 0.346 & 0.602 & 0.643 & 0.411 & 0.422 & 0.395 & 0.693 & 0.678 & 0.458 & 0.453 & 0.584 & 0.544 & 0.094 & 0.139 & 0.145 & 0.439 \\ \hline \hline
\end{tabular}
\end{table*}

\subsection{Uncertainty Rectification Module}
\label{sec-rectification}

When dealing with challenging scenarios like the segmentation of targets characterized by ambiguous boundaries, SAM tends to exhibit a tendency of mis-segmenting unrelated regions primarily associated with higher uncertainty, therefore leading to poor segmentation performance.
To rectify the segmentation result based on estimated uncertainty, \cite{SAM-FNPC} uses a straightforward approach to set a pre-defined threshold $u_{th}$ for selecting high-uncertainty areas.
Then the possible false positive and false negative regions are identified based on the intersection of the high-uncertainty mask with the segmentation mask and background mask respectively.
These potential false negative or false positive regions are then added or removed from the initial segmentation result to obtain the final segmentation result.

However, we argue that this simple 'correction' may not improve the performance or even introduce additional segmentation errors, since regions with high uncertainty may not necessarily correspond to false positives or negatives, especially for complex organs with different shapes and volumes. In some cases, they may be ambiguous or complex regions where the target is unclear or difficult to define.
To this end, we propose to rectify uncertain regions based on the assumptions that 1) pixels that have similar intensity and 2) are close to each other in the image are likely to belong to the same class.
Instead of setting a fixed threshold for the selection of high uncertainty areas, we use a class-specific threshold for confidence-based filtering 
as follows:
\begin{equation}
T_{unc} = min( u(\hat y)) + \frac{ S_{\hat{y}} + S_{b}  }{2 \times S_{b}  } [max( u(\hat y))-min( u(\hat y)) ]
\label{threshold}
\end{equation}
where $S_{\hat{y}}$ and $ S_{b}$ represent the area of the ensembled segmentation mask and corresponding bounding box, respectively.
Lower ratio indicates that the target occupies a smaller portion of the bounding box and there might be more uncertainty in regions surrounding the target including potential false positives or false negatives in the segmentation result.
Then the high-uncertainty areas are selected.
\begin{equation}
M_{unc} = u(\hat y) > T_{unc}
\label{uncmask}
\end{equation}

\begin{algorithm}[t]
	\caption{The overall workflow of our proposed Uncertainty Rectified Segment Anything Model (UR-SAM).}
	\label{Algorithm1}
	\begin{algorithmic}[1]
		  \STATE Input an image $x$ 
		  \STATE Locate extreme points of the target organ and generate bounding box prompt $b$ and corresponding bounding box $S_b$.
            \STATE Augment the prompt $b$ to generate $B$ consists of $\{b^1,b^2,\dots,b^n\}$ based on pre-defined number and ratio.
            \STATE Generate segmentation outputs $\{y^1,y^2,\dots,y^n\}$ based on multi-prompt inputs $B$.
	   	\STATE Generate ensembled segmentation result $\hat y$, ensembled segmentation mask $S_{\hat{y}}$ and   estimated uncertainty $u(\hat y)$ as Eq.(\ref{ensemble})  and Eq.(\ref{unc}).
            \STATE Generate high uncertainty mask $M_{unc}$ with class-specific threshold as Eq.(\ref{threshold}) and Eq.(\ref{uncmask}). 
            \STATE Calculate the average intensity of the image within target area $I_{t}$ and background area $I_{b}$
		  \STATE Initialize $\hat y_{r} = M_{t} = \hat y * (1- M_{unc})$
            \FOR{ $(i,j)$ where $M_{unc}(i,j) = 1$}
                \IF{$ \frac{ I_{t} - I_{b}}{2} < x(i,j) < \alpha_{h} \times I_{t} $}
                     \STATE $\hat y_{r}(i,j)=1$
                \ENDIF
            \ENDFOR
		\RETURN Rectified segmentation result $\hat y_{r}$
	\end{algorithmic}
\end{algorithm}

By confidence-based filtering, we can divide the image into three parts, $M_{t}$ represents the certain regions of target inside high uncertainty areas, $M_{b}$ represents the certain regions of background outside the uncertainty areas, and $M_{unc}$ represents the uncertain regions with high uncertainty.
For rectification of uncertain regions, we estimate the average intensity of the image within certain regions of target and background as $I_{t}$ and $I_{b}$.
If the pixel intensity of $M_{unc}$ is within a certain range of image intensity included in $M_{t}$, it is included to be part of the final segmentation result. The overall workflow is illustrated in \ref{Algorithm1}.

\section{Experiments}

\subsection{Dataset and Model}

We conduct experiments on two different medical image segmentation datasets. The first dataset is Automatic Structure Segmentation for Radiotherapy Planning Challenge Task1 dataset \textbf{(StructSeg)}\footnote{https://structseg2019.grand-challenge.org}, which contains 50 CT scans for the segmentation of 22 head-and-neck (HaN) organs including brain stem (BS), left eye (E-L), right eye (E-R), left lens (L-L), right lens (R-L), left optic nerve (ON-L), right optic nerve (ON-R), optic chiasma (OC), left temporal lobes (TL-L), right temporal lobes (TL-R), pituitary (P), left parotid gland (PG-L), right parotid gland (PG-R), left inner ear (IE-L), right inner ear (IE-R), left middle ear (ME-L), right middle ear (ME-R), left TM joint (J-L), right TM joint (J-R), spinal cord (SC), left mandible (M-L) and right mandible(M-R). 
The second dataset is the labeled set of Fast and Low-resource semi-supervised Abdominal oRgan sEgmentation Challenge \textbf{(FLARE 22)} \cite{FLARE22}, which contains 50 CT scans for the segmentation of 13 abdominal organs including liver, right kidney, spleen, pancreas, aorta, inferior vena cava (IVC), right adrenal gland (RAG), left adrenal gland (LAG), gallbladder (Gall), esophagus, stomach, duodenum, and left kidney.
These two datasets are highly representative as they contain the vast majority of important organs in the human body, providing good coverage for organ structure segmentation tasks in common clinical scenarios. 
To validate the effectiveness of our proposed framework, we integrate two different segmentation models with ViT-B backbone for the experiments: the original SAM \cite{SAM} and MedSAM \cite{MedSAM}, a specialized variant of SAM fine-tuned on medical image datasets.

\subsection{Implementation Details and Evaluation Metrics}
All of our experiments are conducted using an NVIDIA A100 GPU.
Following the settings in \cite{MedLSAM}, we randomly select five scans as support images for both datasets.
For each organ in these scans, we compute the maximum and minimum coordinates and take the average of the coordinates and features across the support images to generate an average representation of latent coordinates and feature extreme points for auto-prompting. Specifically, the voxel spacing of input images is re-scaled to $3\times3\times3$ $mm^{3}$ with cropping patch sizes of $64\times64\times64$ pixels. The 3D bounding box obtained from localization is extended by [2, 10, 10] pixels in the z, x, and y directions, respectively to ensure that the targeted organ is completely encapsulated within the box.
For data preprocessing, we perform the same procedure as described in \cite{MedSAM}, which includes adjusting the slice resolution to 3 × 1024 × 1024 and normalizing all images to the range [-500, 1000] as this range effectively encompasses most tissues.
In addition to the automatic localization to generate prompts, we also conduct a series of experiments for manual prompting-based interactive segmentation, which are simulated based on the ground truth. To mimic the potential inaccuracies in manually drawn bounding boxes, we introduce a random perturbation of 0-20 pixels.
In our experiments, we use the Dice Similarity Coefficient (DSC) as the evaluation metric of the segmentation task, which is one of the most commonly used evaluation metric for image segmentation to measure the degree of pixel-wise overlap between the segmentation results and the ground truth. Higher DSC indicate better segmentation performance. The calculation is defined as follows:

\begin{equation}
DSC(G, S) = \frac{2|G\cap S|}{|G| + |S|}
\label{DSC}
\end{equation}

\begin{table}
	\caption{Ablation analysis of different components of uncertainty rectification approach.} \label{Ablation}
	\centering
\footnotesize
\label{Table_ablation}
\renewcommand\arraystretch{1.3}
\begin{tabular}{c|cc|cc}
\hline \hline
    Dataset                     & \multicolumn{2}{c}{StructSeg DSC} & \multicolumn{2}{c}{FLARE22 DSC} \\ \hline
    Backbone                     &   SAM       &  MedSAM     & SAM       &  MedSAM       \\ \hline
baseline             &     0.394         &     0.453      &   0.527     &   0.448     \\ 
ensemble             &     0.493    &     0.478      &    0.545    &     0.527     \\ \hline
fixed $T_{unc}$ + Rec       &  0.498     &   0.484      &  0.569     &     0.581       \\
class-specific $T_{unc}$ + Rec &     0.501     &    0.486     &   0.572    &    0.586       \\ \hline \hline
\end{tabular}
\end{table}

\begin{table*}
	\caption{Quantitative evaluation of Dice Similarity Coefficient (DSC) of different rectification methods with comparison to auto-prompting and manual prompting settings using SAM / MedSAM backbone for 3D head-and-neck organ segmentation in the StructSeg dataset. Higher values represent better segmentation performance.} \label{UR_StructSeg}
	\centering
 \label{Table_StructSeg}
	\renewcommand\arraystretch{1.2}
        \scriptsize
\begin{tabular}{p{1.2cm}|p{0.3cm}p{0.3cm}p{0.3cm}p{0.3cm}p{0.3cm}p{0.3cm}p{0.3cm}p{0.3cm}p{0.3cm}p{0.3cm}p{0.3cm}p{0.3cm}p{0.3cm}p{0.3cm}p{0.3cm}p{0.3cm}p{0.3cm}p{0.3cm}p{0.3cm}p{0.3cm}p{0.4cm}p{0.4cm}|c}
\hline \hline
SAM Backbone & BS & E-L & E-R & L-L & L-R & ON-L & ON-R & OC & TL-L & TL-R & P & PG-L & PG-R & IE-L & IE-R & ME-L & ME-R & J-L & J-R & SC & M-L & M-R & Avg \\ \hline
Auto Prompting & 0.540 & 0.612 & 0.630 & 0.227 & 0.201 & 0.304 & 0.344 & 0.297 & 0.253 & 0.196 & 0.361 & 0.064 & 0.080 & 0.529 & 0.640 & 0.635 & 0.640 & 0.544 & 0.580 & 0.091 & 0.479 & 0.431 & 0.394 \\ \hline
Ensemble & 0.427 & 0.636 & 0.555 & 0.376 & 0.255 & 0.340 & 0.376 & 0.331 & 0.568 & 0.334 & 0.376 & \textbf{0.557} & 0.360 & 0.575 & \textbf{0.658} & \textbf{0.784} & \textbf{0.735} & 0.613 & 0.652 & 0.262 & \textbf{0.526} & 0.559 & 0.493 \\
Unc-FPC & 0.524 & 0.540 & 0.661 & 0.138 & 0.376 & 0.457 & \textbf{0.493} & 0.459 & 0.425 & \textbf{0.512} & 0.555 & 0.074 & 0.156 & 0.495 & 0.609 & 0.600 & 0.546 & 0.414 & 0.642 & 0.313 & 0.451 & \textbf{0.605} & 0.457 \\
Unc-FNC & \textbf{0.540} & 0.376 & 0.646 & \textbf{0.634} & \textbf{0.459} & 0.494 & 0.411 & \textbf{0.518} & 0.520 & 0.376 & 0.275 & 0.420 & 0.303 & 0.590 & 0.546 & 0.609 & 0.682 & 0.703 & \textbf{0.723} & 0.303 & 0.478 & 0.504 & \textbf{0.505} \\
Unc-FPNC & 0.491 & 0.455 & 0.558 & 0.608 & 0.325 & 0.451 & 0.376 & 0.383 & 0.572 & 0.449 & \textbf{0.598} & 0.054 & 0.044 & 0.359 & 0.559 & 0.550 & 0.528 & \textbf{0.844} & 0.660 & 0.252 & 0.494 & 0.497 & 0.460 \\ 
\textbf{UR-SAM} & 0.510 & \textbf{0.690} & \textbf{0.725} & 0.243 & 0.310 & \textbf{0.589} & 0.487 & 0.381 & \textbf{0.585} & 0.495 & 0.499 & 0.428 & \textbf{0.462} & \textbf{0.616} & 0.472 & 0.539 & 0.633 & 0.590 & 0.506 & \textbf{0.324} & 0.440 & 0.488 & 0.501 \\ \hline 
Manual Prompting & 0.662 & 0.632 & 0.659 & 0.188 & 0.190 & 0.323 & 0.326 & 0.397 & 0.277 & 0.236 & 0.365 & 0.269 & 0.300 & 0.556 & 0.626 & 0.723 & 0.739 & 0.579 & 0.607 & 0.448 & 0.854 & 0.803 & 0.489 \\ \hline \hline
MedSAM Backbone & BS & E-L & E-R & L-L & L-R & ON-L & ON-R & OC & TL-L & TL-R & P & PG-L & PG-R & IE-L & IE-R & ME-L & ME-R & J-L & J-R & SC & M-L & M-R & Avg \\ \hline
Auto Prompting & 0.643 & 0.652 & 0.664 & 0.280 & 0.264 & 0.385 & 0.369 & 0.320 & 0.705 & 0.733 & 0.404 & 0.532 & 0.464 & 0.669 & 0.644 & 0.480 & 0.435 & 0.562 & 0.436 & 0.117 & 0.050 & 0.161 & 0.453 \\  \hline
Ensemble & \textbf{0.686} & 0.696 & 0.663 & 0.285 & 0.265 & 0.396 & 0.365 & 0.359 & \textbf{0.677} & \textbf{0.708} & 0.411 & 0.511 & 0.491 & \textbf{0.714} & \textbf{0.700} & \textbf{0.505} & \textbf{0.498} & \textbf{0.603} & \textbf{0.542} & 0.108 & 0.168 & 0.170 & 0.478 \\
Unc-FPC & 0.564 & 0.503 & 0.542 & 0.154 & 0.186 & \textbf{0.631} & \textbf{0.545} & 0.464 & 0.627 & 0.542 & 0.514 & 0.663 & \textbf{0.641} & 0.592 & 0.648 & 0.499 & 0.316 & 0.393 & 0.431 & 0.234 & \textbf{0.395} & 0.280 & 0.471 \\ 
Unc-FNC & 0.579 & 0.594 & \textbf{0.786} & \textbf{0.390} & \textbf{0.353} & 0.563 & 0.319 & \textbf{0.581} & 0.517 & 0.578 & 0.104 & \textbf{0.708} & 0.555 & 0.659 & 0.579 & 0.477 & 0.456 & 0.610 & 0.450 & \textbf{0.254} & 0.285 & 0.259 & 0.484 \\
Unc-FPNC & 0.498 & 0.558 & 0.576 & 0.339 & 0.330 & 0.564 & 0.233 & 0.335 & 0.563 & 0.562 & \textbf{0.641} & 0.700 & 0.506 & 0.397 & 0.300 & 0.450 & 0.402 & 0.463 & 0.294 & 0.222 & \textbf{0.395} & 0.253 & 0.436 \\ 
\textbf{UR-SAM} & 0.672 & \textbf{0.773} & 0.730 & 0.272 & 0.265 & 0.552 & 0.426 & 0.376 & 0.636 & 0.628 & 0.505 & 0.494 & 0.418 & 0.573 & 0.688 & 0.443 & 0.495 & 0.526 & 0.499 & 0.174 & 0.258 & \textbf{0.287} & \textbf{0.486} \\  \hline 
Manual Prompting & 0.763 & 0.620 & 0.661 & 0.247 & 0.256 & 0.417 & 0.403 & 0.416 & 0.793 & 0.831 & 0.483 & 0.605 & 0.527 & 0.704 & 0.688 & 0.531 & 0.471 & 0.572 & 0.438 & 0.576 & 0.559 & 0.616 & 0.553 \\ \hline \hline
\end{tabular}
\end{table*}

\begin{figure*}[]
    \centering
	\includegraphics[width=16cm]{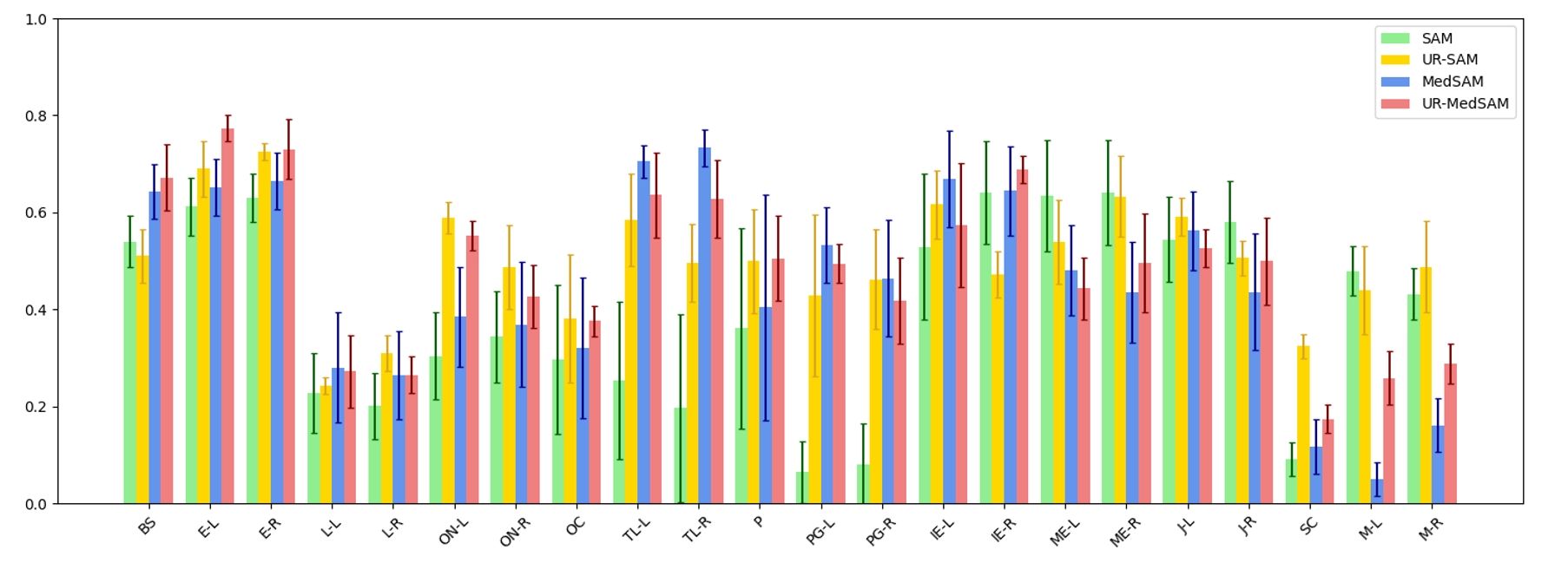}
	\caption{Performance of original and uncertainty rectified segmentation results based on SAM \cite{SAM} and MedSAM \cite{MedSAM} for 3D head-and-neck organ segmentation in the StructSeg dataset.}
	\label{Plot_StructSeg}
\end{figure*}

\subsection{Ablation Analysis}
In this section, we aim to evaluate the effectiveness of our uncertainty rectification approach.
As shown in Table \ref{PromptAug_StructSeg}, we first conduct experiments using different augmentation numbers and perturb ratios for prompt augmentation to select the optimal parameters for uncertainty estimation. From the results, we observe that incorporating perturbations into input prompts can improve the segmentation performance to some extent and enhance the model's overall robustness. However, the benefits of additional augmentations were not observed when the number of augmentations exceeded three, or in some cases, even led to decreased segmentation performance. It is noteworthy that high perturbation ratios may cause some parts of the target to fall outside of the bounding box, resulting in decreased segmentation performance. Therefore, the augmentation ratio should be set in a suitable range.
Table \ref{Table_ablation} presents the ablation analysis of different components of our framework. It can be seen that using class-specific thresholds for rectification increases the segmentation performance in both datasets.

\begin{table*}[t]
	\caption{Quantitative evaluation of Dice Similarity Coefficient (DSC) of different rectification methods with comparison to auto-prompting and manual prompting settings using SAM / MedSAM backbone for 3D abdominal organ segmentation in the FLARE 22 dataset. Higher values represent better segmentation performance.} \label{UR_FLARE}
	\centering
	\renewcommand\arraystretch{1.3}
        \label{Table_FLARE}
        \scriptsize
\begin{tabular}{m{1.15cm}|ccccccccccccc|c}
\hline \hline
SAM    Backbone & Liver & Kidney-R & Spleen & Pancreas & Aorta & IVC & RAG & LAG & Gall & Esophagus & Stomach & Duodenum & Kidney-L  & Avg \\ \hline
Auto Prompting & 0.665 & \textbf{0.850} & 0.668 & 0.345 & 0.489 & \textbf{0.640} & 0.278 & 0.398 & 0.554 & 0.316 & 0.480 & 0.331 & \textbf{0.849} & 0.527 \\ \hline
Ensemble & 0.550 & 0.593 & 0.466 & 0.516 & 0.535 & 0.466 & 0.476 & 0.666 & \textbf{0.773} & 0.641 & 0.457 & 0.379 & 0.572 & 0.545 \\
Unc-FPC & 0.521 & 0.501 & 0.465 & 0.442 & 0.445 & 0.526 & 0.237 & 0.448 & 0.586 & \textbf{0.726} & 0.423 & 0.322 & 0.574 & 0.478 \\
Unc-FNC & 0.510 & 0.556 & \textbf{0.676} & 0.494 & 0.507 & 0.488 & 0.469 & \textbf{0.682} & 0.528 & 0.417 & 0.449 & \textbf{0.451} & 0.689 & 0.532 \\
Unc-FPNC & 0.486 & 0.466 & 0.582 & 0.332 & \textbf{0.597} & 0.284 & \textbf{0.777} & 0.630 & 0.065 & 0.589 & 0.355 & 0.404 & 0.473 & 0.457 \\ 
\textbf{UR-SAM} & \textbf{0.666} & 0.742 & 0.671 & \textbf{0.555} & 0.587 & 0.585 & 0.327 & 0.353 & 0.748 & 0.444 & \textbf{0.608} & 0.376 & 0.776 & \textbf{0.572} \\ \hline
Manual Prompting & 0.799 & 0.956 & 0.928 & 0.707 & 0.912 & 0.879 & 0.527 & 0.643 & 0.826 & 0.707 & 0.831 & 0.561 & 0.952 & 0.787 \\ \hline \hline
MedSAM Backbone & Liver & Kidney-R & Spleen & Pancreas & Aorta & IVC & RAG & LAG & Gall & Esophagus & Stomach & Duodenum & Kidney-L  & Avg  \\ \hline
Auto Prompting & 0.434  & 0.546  & 0.451 & 0.299  & 0.476  & 0.499 & 0.186 & 0.425 & 0.608 & 0.608 & 0.643 & 0.126  & 0.523 & 0.448 \\ \hline
Ensemble & 0.492 & 0.494 & 0.523 & 0.332 & 0.479 & \textbf{0.724} & 0.304 & 0.731 & 0.635 & 0.533 & 0.538 & \textbf{0.581} & 0.491 & 0.527 \\
Unc-FPC & 0.504 & 0.498 & 0.548 & 0.435 & 0.495 & 0.575 & 0.164 & 0.664 & 0.441 & 0.664 & 0.500 & 0.458 & 0.576 & 0.502 \\
Unc-FNC & 0.509 & 0.618 & 0.492 & \textbf{0.558} & 0.535 & 0.628 & \textbf{0.725} & 0.692 & 0.469 & \textbf{0.742} & 0.458 & 0.462 & 0.510 & 0.569 \\
Unc-FPNC & 0.507 & 0.501 & 0.483 & 0.359 & 0.421 & 0.472 & 0.601 & 0.600 & \textbf{0.782} & 0.333 & 0.518 & 0.501 & 0.602 & 0.514 \\ 
\textbf{UR-SAM} & \textbf{0.599} & \textbf{0.653} & \textbf{0.565} & 0.444 & \textbf{0.580} & 0.599 & 0.320 & \textbf{0.752} & 0.661 & 0.680 & \textbf{0.544} & 0.513 & \textbf{0.705} & \textbf{0.586} \\ \hline 
Manual Prompting & 0.685 & 0.724 & 0.782 & 0.784 & 0.714 & 0.736 & 0.466 & 0.449 & 0.651 & 0.618 & 0.638 & 0.376 & 0.693 & 0.640  \\ \hline \hline
\end{tabular}
\end{table*}

\begin{figure*}[]
    \centering
	\includegraphics[width=18cm]{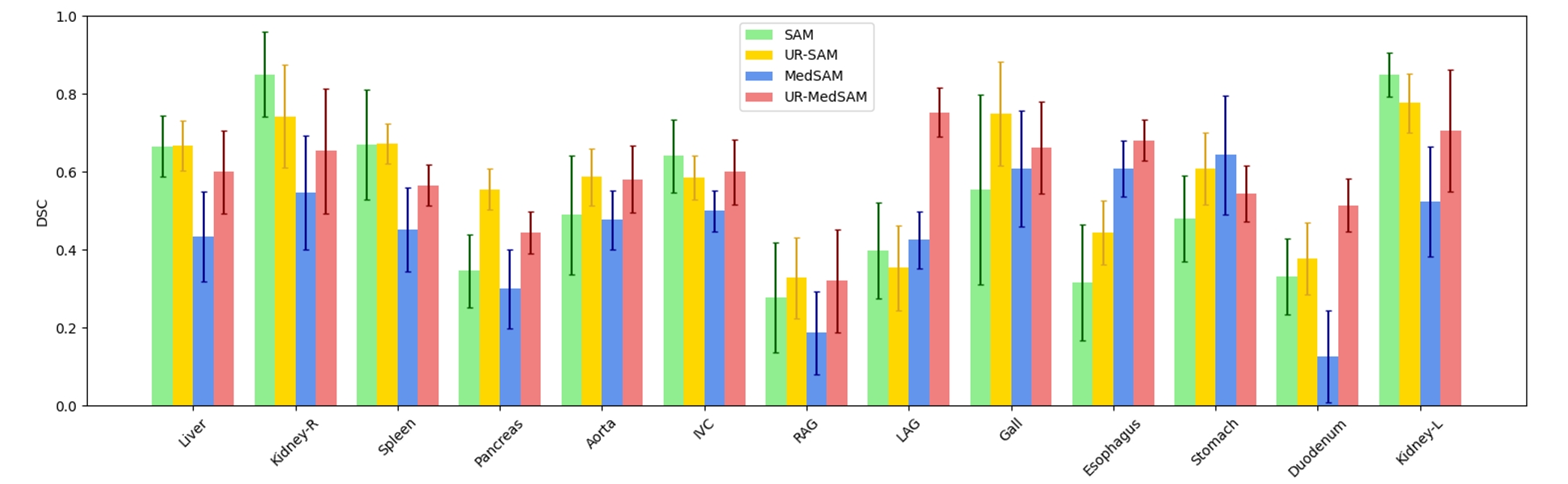}
	\caption{Performance of original and uncertainty rectified segmentation results based on SAM \cite{SAM} and MedSAM \cite{MedSAM} for 3D abdominal organ segmentation in the FLARE 22 dataset.}
	\label{Plot_FLARE}
\end{figure*}

\begin{figure*}[t]
	\includegraphics[width=17.5cm]{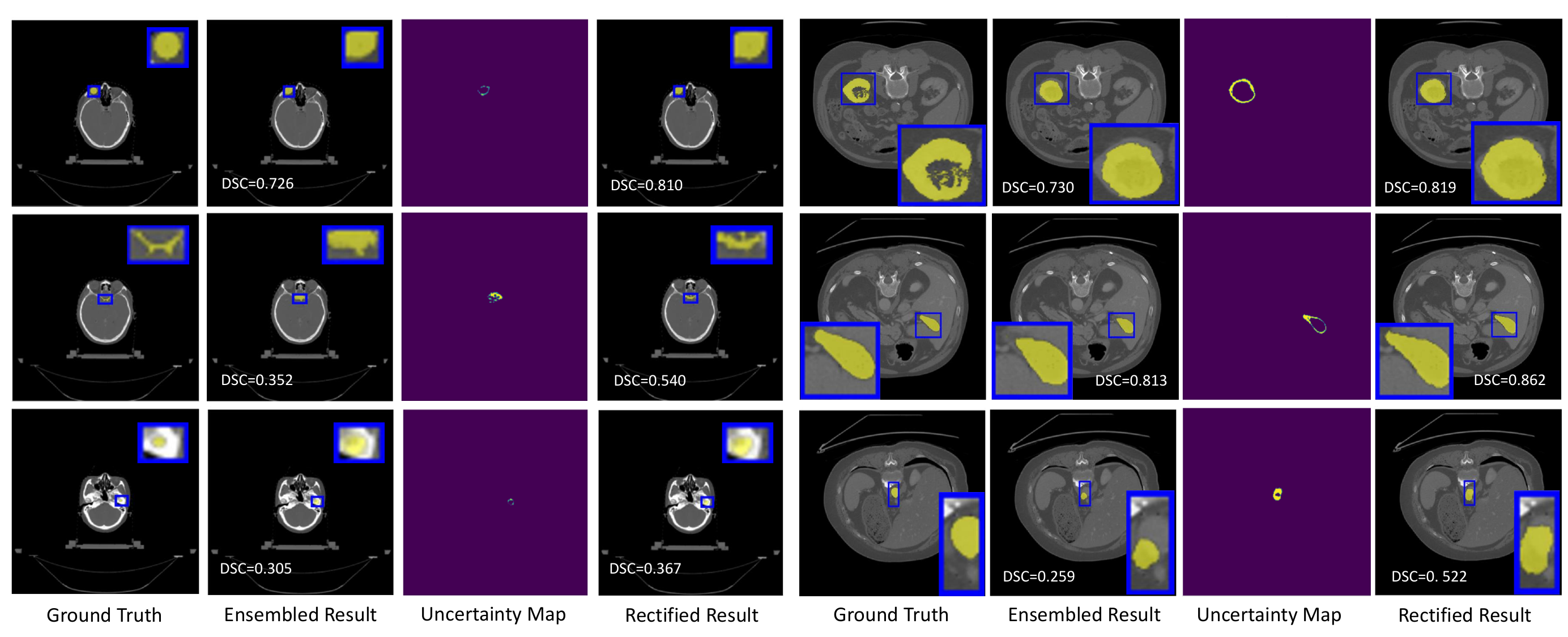}
	\caption{Visual comparison of segmentation results before and after uncertainty rectification on StructSeg and FLARE 22 datasets.}
	\label{Visualize}
\end{figure*}

\begin{table*}[]
	\caption{Comparison of our method with state-of-the-are specialist segmentation models under the same experimental settings. Specifically, N denotes the count of slices containing the target object.} \label{Table_sup}
	\centering
\small
\renewcommand\arraystretch{1.5}
\begin{tabular}{c|cc|c}
\hline \hline
    Methods               & Manual Annotation & Manual Prompting  & StructSeg DSC  \\ \hline
    nnUNet-2D \cite{isensee2020nnunet}  &  5 labeled images  & - & 0.422   \\ 
    nnUNet-3D \cite{isensee2020nnunet} &  5 labeled images  & -  & 0.557   \\ \hline
    SAM \cite{SAM}      &  -  & N bboxes for each target &  0.489  \\
    MedSAM \cite{MedSAM} & - & N bboxes for each target & 0.553  \\ \hline
    MedLSAM (SAM backbone) \cite{MedLSAM}  &  5 support images & - &  0.394 \\
    MedLSAM (MedSAM backbone) \cite{MedLSAM}  &  5 support images & - &  0.453  \\ \hline
    \textbf{UR-SAM (SAM backbone)} &   5 support images & - &  0.501     \\ 
    \textbf{UR-SAM (MedSAM backbone)} &   5 support images & - &  0.486      \\  \hline \hline
\end{tabular}
\end{table*}

\subsection{Comparison Experiments}

We first conduct experiments on auto-prompting setting where the landmark detection model introduced in Sec. \ref{Landmark} is utilized to detect extreme points and generate bounding boxes as prompts for segmentation.
We compare our rectification method with the FN/FP correction strategy in \cite{SAM-FNPC}, where high-uncertainty areas are assumed to be potential false negative and positive regions for correction by removing potential false positive regions (Unc-FPC), adding potential false negative regions (Unc-FNC) and correcting both potential false negative and positive regions (Unc-FPNC) from the initial segmentation result.
In addition to the auto-prompting, we also include manual prompting simulated based on the ground-truth masks for comparison.

The experimental results of 3D head-and-neck organ segmentation in shown in Table \ref{Table_StructSeg}.
The head-and-neck region typically comprises relatively small organs with irregular shapes. 
We observe that directly applying SAM fails to segment target organs like the spinal cord (SC) and parotid gland (PG-L / PG-R), with an average dice coefficient lower than 10\%, while ensembled results with augmented prompts can significantly improve the performance of these targets.
For these small organs, the precise location of region of interest is crucial for the accuracy of subsequent segmentation and prompt augmentation can somehow ease the negative influence of generated inaccurate prompts. 
For 3D abdominal organ segmentation in Table \ref{Table_FLARE}, most target organs can be successfully segmented when directly applying SAM, especially for organs with neat shapes and well-defined boundaries like the liver and kidneys. 
However, we observe that prompt augmentation cannot consistently enhance performance and may even decrease the performance for some targets, since part of the organ may extend beyond the augmented bounding box in some instances.

For rectification results, we observe that the performance of comparing strategies varied for different target organs due to the complex shape differences.
Given that regions with high uncertainty may not necessarily correspond to false positives or negatives, especially for complex organs with differing shapes and volumes, this approach did not consistently improve performance. In comparison, our rectification strategy demonstrates more robust improvements for most segmentation targets, resulting in overall superior performance.
We compare the performance of original and uncertainty rectified segmentation results in Fig. \ref{Plot_StructSeg} and \ref{Plot_FLARE}. It can be observed that our framework further improves the segmentation performance with up to 10.7 \% for head-and-neck organ segmentation and 13.8 \% for abdominal organ segmentation in average dice coefficient. Fig. \ref{Visualize} presents the visual analysis of our proposed framework. The four columns represent the ground truth, ensembled segmentation result, high uncertainty mask, and rectified segmentation result, respectively.
We can observe that rectified segmentation results have superior performance with less segmentation error compared with ensembled segmentation results, demonstrating the effectiveness of our proposed framework.

In Table \ref{Table_sup}, we make a comparative analysis of SAM-based auto-prompting segmentation methods and interactive segmentation methods based on manual prompting with training-based state-of-the-art automatic segmentation methods like nnUNet \cite{isensee2020nnunet}.
Despite achieving zero-shot segmentation without the need for voxel-wise annotations, both the original SAM \cite{SAM} and its medical adaptions \cite{MedSAM} require manual prompting of each target organ for each testing case to obtain the segmentation, which directly increase the burden for applications.
Compared with training-based task-specific segmentation methods, SAM-based auto-prompting methods can leverage the knowledge base of foundation models for automatic segmentation, requiring training only a lightweight localization model. For instance, the training of 3D nnUNet needs around $\sim$ 25 hours, while the localization model only $\sim$ 30 minutes.

\section{Discussion and Conclusion}

In this work, we present UR-SAM, an uncertainty rectified SAM framework for auto-prompting medical image segmentation by leveraging prompt augmentation of generated bounding box prompts for uncertainty evaluation and utilizing estimated uncertainty for rectification of segmentation results to improve the accuracy. 
Furthermore, the uncertainty map can help identify potential segmentation errors and support further analysis, offering valuable guidance in areas where manual focus and refinement are required for clinicians. The framework could also be integrated into interactive mechanisms, allowing users to select and modify specific regions that require refinement.
Besides, we observe that MedSAM underperforms SAM to some extent, which is in line with the observations in \cite{MedLSAM}.
This indicates that despite being fine-tuned on relatively large-scale medical datasets, the model does not consistently surpass classic SAM when applied to previously unseen datasets.

While our method has demonstrated significant improvements, there are aspects where further enhancements could be made. 
One notable limitation is that we use a relatively simple strategy to classify pixels in regions with high uncertainty, which relied solely on the intensity of the pixel and did not consider information from neighboring pixels. Consequently, it still yields unsatisfactory results when it comes to segmenting some challenging classes.
Future work will focus on incorporating contrast learning of pixels between different classes for more comprehensive evaluations. Besides, it would be interesting to utilize uncertainty in the training procedure to select out more representative regions to further enhance model performance \cite{wang2022uncertainty,tang2022unified}.


\bibliographystyle{IEEEtran}
\bibliography{ref}

\end{document}